\documentclass[11pt]{article}

\usepackage[preprint]{acl}

\usepackage{times}
\usepackage{latexsym}
\usepackage{amsfonts}

\usepackage{booktabs}
\usepackage{amsmath}
\usepackage{xcolor}
\usepackage{soul}

\usepackage{subcaption}

\usepackage[T1]{fontenc}

\usepackage[utf8]{inputenc}

\usepackage{microtype}

\usepackage{inconsolata}

\usepackage{graphicx}

\title{Sentence-Anchored Gist Compression for Long-Context LLMs}

\author{Dmitrii Tarasov \\
        FusionBrainLab \\
        HSE University \\
        {\small\texttt{tarasov@fusionbrainlab.com}} \\\And
  Elizaveta Goncharova \\
        FusionBrainLab \\
        HSE University \\\And
  Kuznetsov Andrey \\
          FusionBrainLab \\
          Innopolis University \\
}

\begin{document}
\maketitle

\begin{abstract}
This work investigates context compression for Large Language Models (LLMs) using learned compression tokens to reduce the memory and computational demands of processing long sequences. We demonstrate that pre-trained LLMs can be fine-tuned to compress their context by factors of 2x to 8x without significant performance degradation, as evaluated on both short-context and long-context benchmarks. Furthermore, in experiments on a 3‑billion‑parameter LLaMA model, our method achieves results on par with alternative compression techniques while attaining higher compression ratios.
\end{abstract}

\section{Introduction}

The growing demand for processing long-context inputs in Large Language Models (LLMs) is often bottlenecked by memory and computational requirements of the self-attention mechanism. In a recent paper \citet{kuratov2025cramming1568tokenssingle} showed that LLMs possess a significant capacity for compression, with an 8B model capable of condensing up to 1568 tokens into a single vector in a prompt-tuning fashion (smaller models achieve lower rates, e.g., ~128 tokens per vector).

We argue that this inherent ability can be extended beyond prompt-tuning. Instead of a single compression token, we propose learning a set of \textit{compression tokens} that pool the most salient information from distinct text segments. These tokens then serve as the conditioning context for subsequent segments during language modeling, effectively creating an information bottleneck, where each segment primarily attends to the compressed representation of the previous one. This architecture can be implemented with minimal modifications to the standard transformer: by extending the model's vocabulary to include the new compression tokens and adjusting the attention mask to enforce the segment-conditioning pattern. This stands in contrast to methods like \citet{activation_beacon} proposed, which, while powerful, originally required Backpropagation Through Time (BPTT) — though a mask-based implementation is also feasible. A key advantage of our setup is that the model learns to compress the context end-to-end using only the standard Language Modeling (LM) objective, eliminating the need for an auxiliary reconstruction loss as proposed by \citet{deng2024gistcompression} to improve uniform gist-based compression.

Furthermore, while preceding works typically employ a uniform distribution of compression tokens \citep{activation_beacon, deng2024gistcompression}, we introduce a rule-based strategy for token insertion at the end of each sentence. This data-dependent positioning aims to align compression boundaries with natural semantic units, thereby facilitating more meaningful and coherent information aggregation.

In summary, our contributions are as follows:
\begin{itemize}
    \item We propose a novel, rule-based strategy for positioning compression tokens to enable data-dependent and semantically-aware context compression.
    \item We implement the proposed method via a simple attention mask modification, which enables efficient parallel processing during both training and prefilling, enhancing scalability.
    \item We extensively evaluate the proposed framework and demonstrate that it maintains strong performance with no significant degradation on both short- and long-context benchmarks.
    \item We achieve high KV-cache compression rates, ranging from 2x to 8x, across various evaluation benchmarks.
\end{itemize}

\section{Related Work}

\paragraph{Sparse Attention.} Sparse attention methods aim to reduce computational costs by limiting the number of tokens to which each token can attend. In Native Sparse Attention \citet{yuan-etal-2025-deepseek-sparse-attention} proposed a hardware-aligned hierarchical sparse attention that combines coarse-grained token compression with fine-grained token selection. The Forgetting Transformer \citep{lin2025adaptive} learns to limit local attention spans via precomputed forgetting scores. 

\paragraph{Recurrence.} Recurrent methods compress context into a fixed state for long-range dependency modeling. The Recurrent Memory Transformer \citep{bulatov2022rmt} learns memory embeddings in an RNN-like fashion, significantly reducing memory requirements. AutoCompressors \citep{chevalier-etal-2023-auto-compressors} extend this idea by learning to generate compression tokens that are used as soft prompts. More recently, the In-context AutoEncoder \citep{ge2024icae} uses an autoregressive encoder to compress context into learned tokens for a decoder, achieving up to 4x compression without BPTT. While effective, many recurrent approaches require Backpropagation Through Time, which can slow training and hinder parallelization.

\paragraph{KV-Cache Compression.} These methods reduce the memory footprint of the Key-Value cache during generation. \citet{R_KV} uses redundancy estimation and importance scoring to retain approximately 10\% of the cache. \citet{li2024snapkv} employs clustering of attention patterns for an 8.2x memory reduction. \citet{zhang2023h2o} evicts tokens based on attention scores, reducing the cache by up to 5x. \citet{wang2024kv_merger} merges cache entries via clustering, achieving 3x compression. FastGen \citep{ge2024model} profiles and applies head-specific strategies (e.g., punctuation, locality) to halve the cache budget. StreamingLLM \citep{xiao2023StreamingLLM} uses attention sinks and a sliding window (~1000 tokens) for streaming applications. SepLLM \citep{chen2024sepllm} extends this by caching separator tokens (e.g., punctuation), showing it generalizes StreamingLLM and improves performance, though it requires task-specific window tuning.

\paragraph{Gist Token Compression.} This line of work introduces learned ``gist'' or ``beacon'' tokens to summarize context. The most related to ours is \citet{activation_beacon}, which processes context in chunks with interleaved beacon tokens to accumulate information. In contrast, our method requires only a single forward pass for the prefill stage and employs data-dependent compression at sentence boundaries. \citet{mu2023gisting} compresses task-specific prompts into gist tokens via a modified attention mask, achieving up to 26x compression. \citet{deng2024gistcompression} explores different compression rates and token placement strategies, using an autoencoding loss to improve compression quality.


\section{Sentence Transformer}

The proposed Sentence transformer takes inspiration from gist or beacon token compression \citep{activation_beacon,deng2024gistcompression} and data-dependent compression from SepLLM \citep{chen2024sepllm}.

\subsection{Learned Gist Tokens}

We extend the LLM's embedding vocabulary by adding $N_g$ new gist tokens. These tokens are initialized by sampling from a multivariate normal distribution whose parameters ($\mu$, $\Sigma$) are derived from the existing vocabulary embeddings, following the ``mean-resizing'' approach by \citet{hewitt2021initializing}. The language modeling head is resized correspondingly. When the LM head is tied to the input embeddings (as is common in smaller models (e.g., 1–3B parameters)) we resize the shared matrix jointly; otherwise, the two matrices are resized separately.

\subsection{Gist Tokens Placement}

The gist tokens in this work serve as aggregators for some text segments. This work employs a simple, rule-based strategy for positioning compression tokens. Specifically, gist tokens are inserted at the end of each sentence, anchored to standard sentence-ending punctuation marks such as `.', `!', and `?'. This approach aims to align compression boundaries with natural semantic units.

\noindent\textbf{Processing Example For $N_g=2$ gist tokens:}
\begin{description}
   \item[Original:] \texttt{The sun was shining brightly. Birds were singing in the forest.}
   \item[Processed:] \texttt{The sun was shining brightly. {\color{red}<$g_1$><$g_2$>}} \texttt{Birds were singing in the forest. {\color{red}<$g_1$><$g_2$>}}
\end{description}

\subsection{Attention Mask}

Within the transformer architecture, the gist tokens for a given sentence are allowed to attend to all tokens within that same sentence, as well as to all gist tokens from preceding sentences. This enables the aggregation of both local sentence-level information and global context from the compressed history. Figure~\ref{fig:sentence_attention} illustrates the sentence attention pattern, which exhibits sparsity despite an increased sequence length. 
This expansion occurs because gist tokens are added to the initial sequence.

\begin{figure}[tbp]
    \centering
    \begin{subfigure}{0.2\textwidth}
        \centering
        \includegraphics[width=\linewidth]{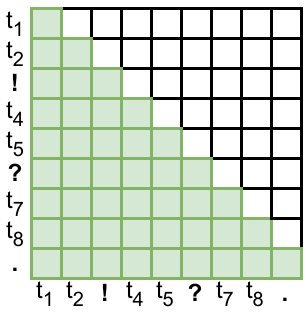}
        \caption{Causal attention}
        \label{fig:causal_attention}
    \end{subfigure}
    \hspace{0.01\textwidth}
    \begin{subfigure}{0.26\textwidth}
        \centering
        \includegraphics[width=\linewidth]{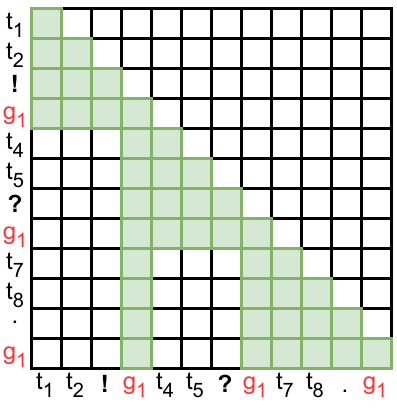}
        \caption{Sentence attention}
        \label{fig:sentence_attention}
    \end{subfigure}
    \caption{Comparison of attention mechanisms: (a)~causal attention, and (b)~sentence attention with $N_g=1$ gist token. 
Gist tokens ({\color{red}$g_1$}) are inserted at sentence boundaries and pool information from their entire sentence. 
They are visible to all subsequent tokens, while regular tokens ($t_i$) only attend within their sentence.}
    \label{fig:sentence_attention}
\end{figure}

\subsection{Objective}

The training methodology relies exclusively on the conventional language modeling objective. Given an input sequence $X = \{x_1, x_2, ..., x_T\}$ and compressed context representation $C$, we optimize:

\begin{equation}
\mathcal{L} = -\mathbb{E}_{x \sim \mathcal{D}} \left[ \sum_{t=1}^{T} \log P(x_t | x_{<t}, C) \right]
\end{equation}

\noindent where $C = f_\theta(X)$ is the compressed context produced by our model parameters $\theta$. This streamlined approach removes the dependency on additional reconstruction losses used by \citet{deng2024gistcompression} to optimize uniform gist compression.

\subsection{Training Stages}\label{sec:training_stages}

Our experimental results demonstrated that initializing training with gist token optimization alone yields superior benchmark performance compared to end-to-end training from the outset. This finding led us to adopt a structured three-phase training approach:

\paragraph{Stage 1: Gist Token Warm-up}
In this initial phase, we freeze all base model parameters and train only the newly introduced gist tokens. This allows the model to learn somewhat context-compression patterns without interfering with the pre-trained representations.

\paragraph{Stage 2: Full Model Fine-tuning}
Next, we unfreeze all parameters and conduct standard fine-tuning across the entire architecture. This stage leverages the initialized compression behavior from Stage 1 while adapting the base model to effectively utilize the compressed representations.

\paragraph{Stage 3: Large-Batch Cold Down}
To further enhance model convergence and stability, we implement a final training stage with significantly increased batch sizes up to 1M tokens with linear decay learning rate schedule following \citet{zhai2021colddown}. 

For our primary experiments, we utilized \texttt{Llama3.2-3B} \citep{grattafiori2024llama3herdmodels} as the base architecture, applying this staged training protocol across multiple downstream tasks and evaluation benchmarks.

\section{Experiments}

For \texttt{Llama3.2-3B} \citep{grattafiori2024llama3herdmodels} model we trained 4 checkpoints with corresponding $1$, $2$, $4$, $8$ gist tokens compression. 

For all stages we used randomly sampled subsets of FineWeb-Edu \citep{lozhkov2024fineweb-edu} total tokens budget for all training stages is approximately $4$B tokens. Other training details can be found in Appendix~\ref{sec:app_training_details}.

\subsection{Short-Context benchmarks}
First, we evaluate the model on short-context benchmarks: HellaSwag \citep{zellers-etal-2019-hellaswag}, ARC \citep{Clark2018ThinkYH_ARC}, MMLU \citep{hendrycks2021measuring}, and WinoGrande \citep{sakaguchi2019winogrande}. These benchmarks probe general linguistic and academic abilities, allowing us to assess whether short-context compression affects performance. During Stage 2 of training (Section~\ref{sec:training_stages}), accuracy on these tasks saturates quickly and closely matches that of the vanilla base model. Results are reported in Table~\ref{tab:short_bench_results}.

On short benchmarks, compression rates range from 16.55 for \texttt{HellaSwag} to 82.96 for \texttt{WinoGrande} with $N_g=1$. 
These rates decrease by approximately half with each doubling of $N_g$. Detailed compression rates across benchmarks are provided in Appendix~\ref{sec:app_bench_compression_rates}.

\subsection{Long-Context Benchmarks}

For long-context evaluation, we employed the HELMET benchmark \citep{yen2025helmet}. To enable rapid evaluation during training, we created a reduced version, \texttt{HELMET (Tiny)}, consisting of 100 samples per task. The performance results on this benchmark are presented in Table~\ref{tab:sep_cache_beacon_comparison}. For most tasks, performance on \texttt{HELMET (Tiny)} improved steadily as the number of gist tokens ($N_g$) increased. Also our model maintained performance comparable to the baseline. We compared our model with strong 7B models: SepLLM \citep{chen2024sepllm} (training-free method) and activation beacon \citep{activation_beacon}, the most close to ours work. Despite being half the size of the compared models, our method achieves performance comparable strong baselines while attaining higher compression rates. For Sentence Llama with $N_g=4$, the average compression rate across all long-context benchmarks is approximately 6$\times$ (see Appendix~\ref{sec:app_bench_compression_rates} for task-specific rates). This compares favorably with the Activation Beacon model, which achieves only 2$\times$ KV-cache compression. The complete set of evaluation metrics for HELMET (Tiny) is provided in Appendix~\ref{sec:app_training_stages_evaluation}.

\begin{table}[]
    \centering
    \begin{tabular}{lllllll}
    \hline
     \textbf{Model}   &  \textbf{ARC}   & \textbf{HS}   & \textbf{MMLU}   & \textbf{WG}    \\
    \hline

\multicolumn{5}{l}{\textbf{Llama-3.2-3B Base}} \\
Vanilla      & 59.21 & 70.90       & 42.50        & 65.11 \\ 
\midrule
\multicolumn{5}{l}{\textbf{Sentence Llama-3.2-3B Base}} \\
$N_g=1$     & 49.77 & 63.63       & 38.25        & 58.96   \\
$N_g=2$     & 53.64 & 66.61       & 38.90        & 61.01  \\
$N_g=4$     & 54.59 & 68.07       & 39.73        & 61.48  \\
$N_g=8$     & 55.17 & 67.86       & 39.89        & 61.17  \\
 
    \hline
    \end{tabular}
    \caption{Evaluation results on short-context benchmarks for the Llama3.2-3B model. $N_g$ denotes the number of gist tokens per sentence. HS - HellaSwag, WG - WinoGrande. MMLU cloze.}
    \label{tab:short_bench_results}
\end{table}

\begin{table}[]
    \centering
    \begin{tabular}{llllll}
    \hline
\textbf{Model}           & \textbf{recall}   & \textbf{icl}   & \textbf{longqa}  & \textbf{cite} 
\\
\midrule

\multicolumn{5}{l}{\textbf{Llama-3.2-3B Base}} \\
 Vanilla & 100.0    & 68.2  & 37.3    & 30.0    \\
\midrule
 
 \multicolumn{5}{l}{\textbf{Sentence Llama-3.2-3B Base}} \\
$N_g=1$  & 1.0      & 28.2  & 36.5    & 17.6    \\
$N_g=2$  & 30.0     & 67.4  & 31.8    & 17.7    \\
$N_g=4$  & 90.0     & 69.6  & 32.0    & 20.4    \\
$N_g=8$  & 95.0     & 63.0  & 34.5    & 22.4    \\

\midrule

\multicolumn{5}{l}{\textbf{Llama-3.1-8B-Instruct}} \\

Vanilla  & 100.0     & 15.0 & 29.3      & 34.9   \\
SepCache  & 10.0      & 15.8 & 26.7      & 23.5   \\

\midrule
\multicolumn{5}{l}{\textbf{Qwen2-7B-Instruct}} \\
Vanilla  & 100.0     & 71.2 & 31.4      & 37.4   \\
Beacon  & 88.0      & 64.2 & 27.4      & 32.1   \\

    \hline
    \end{tabular}
    \caption{Sentence Attention Comparison on HELMET (tiny) with SepCache~\citep{chen2024sepllm} and Beacon Compression~\citep{activation_beacon}. $N_g$ denotes the number of gist tokens per sentence.}
    \label{tab:sep_cache_beacon_comparison}
\end{table}

\subsection{Punctuation Sensitivity}\label{sec:punkt_sensitivity}

\sethlcolor{green}

Our analysis revealed that the model's performance is sensitive to punctuation in benchmark templates. The \texttt{icl} benchmark exhibited the most pronounced sensitivity, where the addition of a single period (`.') nearly doubled the measured performance. Without this final period, the model incorrectly compresses a question's label into gist tokens associated with the subsequent question, thereby degrading task performance.

\noindent\textbf{HELMET ICL template modification:}
\begin{description}

    \item[Original:]
    \texttt{\\
        What is swap math ?\\
        label: 4\\
        When does the average teenager first have intercourse ?\\
        label: 5\\
        ...}
    
    \item[With extra dot:] 
    \texttt{\\
        What is swap math ?\\
        label: 4\hl{.}\\
        When does the average teenager first have intercourse ?\\
        label: 5\hl{.}\\
        ...}
\end{description}

\section{Limitations}

Despite achieving significant compression rates on long-context benchmarks, our approach has several limitations that warrant discussion:

\begin{itemize}
    \item \textbf{Rule-based token placement:} The current method relies on rule-based gist token positioning. Future work should explore learned compression token locations for more adaptive context aggregation.
    
    \item \textbf{Fixed compression budget:} Using a fixed number of gist tokens per segment limits flexibility. Dynamic allocation of compression tokens based on content complexity would be more efficient.
    
    
    \item \textbf{Punctuation sensitivity:} As discussed in Section~\ref{sec:punkt_sensitivity}, the rule-based placement strategy introduces sensitivity to punctuation variations in benchmark templates.

    \item \textbf{Performance gap:} The compressed model does not fully recover the performance of the base model without compression.

    \item \textbf{Implementation inefficiency}. The current implementation materializes the full attention mask, which becomes memory-prohibitive for very long contexts (e.g., 128K tokens).

    \item \textbf{Limited model size.} All ablation studies and experiments were conducted on a 3B-parameter model; scaling to larger models is left to future work.

\end{itemize}

\bibliography{custom}

\appendix

\section{Training Details}\label{sec:app_training_details}

All training stages employed the AdamW optimizer \citep{loshchilov2018adamw} with weight decay $0.1$, betas $(\beta_1=0.9, \beta_2=0.95)$, and epsilon $1\times10^{-8}$. Training was conducted in bfloat16 precision on a single node with 8 NVIDIA A100 GPUs. Table~\ref{tab:training_hp} summarizes the hyperparameters used across different training stages, including processed tokens, batch size, optimization steps, learning rate, warmup steps, learning rate schedule type, gradient clipping threshold, and training duration.

\begin{table}
\centering
\begin{tabular}{lccc}
\toprule
 & \textbf{Stage 1} & \textbf{Stage 2} & \textbf{Stage 3} \\
\midrule
Max Seq Len & 4096 & 4096 & 4096 \\
Tokens & 0.1B & 2B & 2B \\
Batch Size & 64 & 128 & 512 \\
Training Time & 3 hours & 30 hours & 30 hours \\
\\
Optim. Steps & 1000 & 9000 & 2000 \\
Max LR & $1\mathrm{e}{-4}$ & $1\mathrm{e}{-4}$ & $5\mathrm{e}{-5}$ \\
Min LR & $5\mathrm{e}{-5}$ & $5\mathrm{e}{-5}$ & $0$ \\
Warmup Steps & 100 & 1000 & 100 \\
LR Schedule & Cosine & Cosine & Linear \\
MaxGradNorm & 1.0 & 2.0 & 2.0 \\
\bottomrule
\end{tabular}
\caption{Training hyperparameters across different stages.}
\label{tab:training_hp}
\end{table}

\section{Evaluation Across Training Stages}\label{sec:app_training_stages_evaluation}

Table~\ref{tab:short_bench_results_full} presents the evaluation results on short-context benchmarks across all training stages. We observe rapid performance saturation, with only marginal improvements as the number of gist tokens ($N_g$) increases.

Table~\ref{tab:helmet_tiny_full} presents the evaluation results on the HELMET (Tiny) benchmark. The initial training stage resulted in performance degradation across most tasks, while the second stage yielded substantial improvements. The third stage provided only marginal additional gains. Notably, on the \texttt{rerank} task, our Sentence Attention model with $N_g=8$ significantly outperformed the base model, achieving an NDCG@10 of 7.32 compared to 0.40 for the base Llama3.2 3B model.

\begin{table}
    \centering
    \begin{tabular}{lllllll}
    \hline
     $N_g$      & \textbf{ARC}   & \textbf{HS}   & \textbf{MMLU}   & \textbf{WG}   & \textbf{PG19}   \\
    \hline

\multicolumn{6}{l}{\textbf{Base}} \\
-      &  59.21 & 70.90       & 42.50        & 65.11        & 9.64   \\
\midrule

\multicolumn{6}{l}{\textbf{Stage 1. Gist Embeddings only}} \\
 1      &  37.65 & 55.59       & 33.06        & 57.46        & 27.06  \\
 2      &  40.94 & 58.85       & 36.45        & 57.77        & 19.24  \\
 4      &  47.82 & 61.70       & 36.55        & 56.99        & 15.68  \\
 8      &  51.42 & 61.77       & 37.37        & 58.01        & 10.48  \\
\midrule

\multicolumn{6}{l}{\textbf{Stage 2. Finetune}} \\
1      &  47.16 & 62.65       & 37.40        & 58.96        & 13.65  \\
 2      &  51.46 & 65.75       & 38.08        & 59.51        & 11.98  \\
 4      &  54.64 & 67.34       & 39.45        & 60.93        & 9.52   \\
 8      &  53.01 & 67.11       & 38.39        & 61.25        & 7.58   \\
\midrule

\multicolumn{6}{l}{\textbf{Stage 3. Cold down}} \\
1      &  49.77 & 63.63       & 38.25        & 58.96        & 12.83  \\
 2      &  53.64 & 66.61       & 38.90        & 61.01        & 11.29  \\
 4      &  54.59 & 68.07       & 39.73        & 61.48        & 9.24   \\
 8      &  55.17 & 67.86       & 39.89        & 61.17        & 7.17   \\

    \hline
    \end{tabular}
    \caption{Short benchmarks evaluation across training stages. Evaluation results on short-context benchmarks for the Llama3.2-3B model. $N_g$ denotes the number of gist tokens per sentence. HS - HellaSwag, WG - WinoGrande, MMLU cloze version was evaluated}
    \label{tab:short_bench_results_full}
\end{table}

\begin{table}
    \centering
    \begin{tabular}{llllllll}
    \hline
    $N_g$                  & \textbf{recall}   & \textbf{rerank}   & \textbf{cite}   & \textbf{longqa}    & \textbf{icl}   \\
    \hline
    
    \multicolumn{6}{l}{\textbf{Vanilla}} \\
    -      & 100   & 0.40     & 30.02  & 37.34    & 68.20 \\
    
    \\
    \multicolumn{6}{l}{\textbf{Stage 1. Gist Embeddings only}} \\
    1      & 0.00     & 0.00     & 13.13  & 12.93       & 2.00  \\
    2      & 0.00     & 0.00     & 15.50  & 20.72       & 0.00  \\
    4      & 0.00     & 0.00     & 17.57  & 23.64       & 0.00  \\
    8      & 0.00     & 0.12     & 20.64  & 30.55       & 15.40 \\
    
    \\
    \multicolumn{6}{l}{\textbf{Stage 2. Finetune}} \\
    1      & 1.00     & 0.00     & 17.56  & 36.50       & 28.20 \\
    2      & 30.00    & 2.34     & 17.74  & 31.81       & 67.40 \\
    4      & 90.00    & 2.05     & 20.43  & 31.96       & 69.60 \\
    8      & 95.00    & 5.80     & 22.38  & 34.53       & 63.00 \\
    
    \\
    
    \multicolumn{6}{l}{\textbf{Stage 3. Finetune}} \\
    1      & 3.00     & 0.18     & 17.87  & 33.94       & 32.80 \\
    2      & 46.00    & 2.00     & 18.09  & 34.45       & 71.20 \\
    4      & 96.00    & 3.06     & 20.86  & 33.36       & 63.20 \\
    8      & 98.00    & 7.31     & 22.94  & 33.49       & 64.60 \\

    \hline
    \end{tabular}
    \caption{HELMET Tiny evaluation results across training stages.}
    \label{tab:helmet_tiny_full}
\end{table}

\section{Evaluation Details}

\subsection{Short-Context Benchmarks}

For the short-context benchmarks evaluation we used lighteval \citep{lighteval} with the same configuration as in SmolLM2 \citep{allal2025smollm2}. For MMLU we used cloze formatting and chose the option with minimal perplexity. All other benchmarks were also evaluated based on the perplexity of the correct answers. There were no generative benchmarks in this subset.

\subsection{HELMET Tiny}

All long-context benchmarks in the HELMET evaluation suite are generative: the model produces a limited number of tokens, and the generated output is then compared against a ground-truth answer. For the HELMET Tiny subset, we selected a small number of samples per task, with sequence lengths ranging from 4K to 8K tokens. Table~\ref{tab:helmet_tiny_description} provides detailed information for each task, including maximum sequence length and sample count.

\begin{table}[htbp]
\centering
\begin{tabular}{lrr}
\toprule
\textbf{Task} & \textbf{Max Length} & \textbf{Samples} \\
\midrule
recall & 4k & 100 \\
rerank & 8k & 100 \\
cite & 8k & 100 \\
longqa & 8k & 100 \\
icl & 8k & 500 \\
\bottomrule
\end{tabular}
\caption{HELMET Tiny benchmark configuration: task specifications and dataset statistics.}
\label{tab:helmet_tiny_description}
\end{table}

\subsection{PG19 Perplexity Evaluation} \label{sec:app_pg19_evaluation}

On PG19 \citep{Rae2020CompressiveTransformerPG19}, higher compression yields lower overall perplexity, but this is driven by gist tokens. When excluding all but the final gist token per segment, perplexity increases significantly. This occurs because gist tokens have predictable patterns, while the final gist token must predict subsequent sentence beginnings, resulting in higher perplexity.

\begin{figure}[h]
    \centering
    \includegraphics[width=\linewidth]{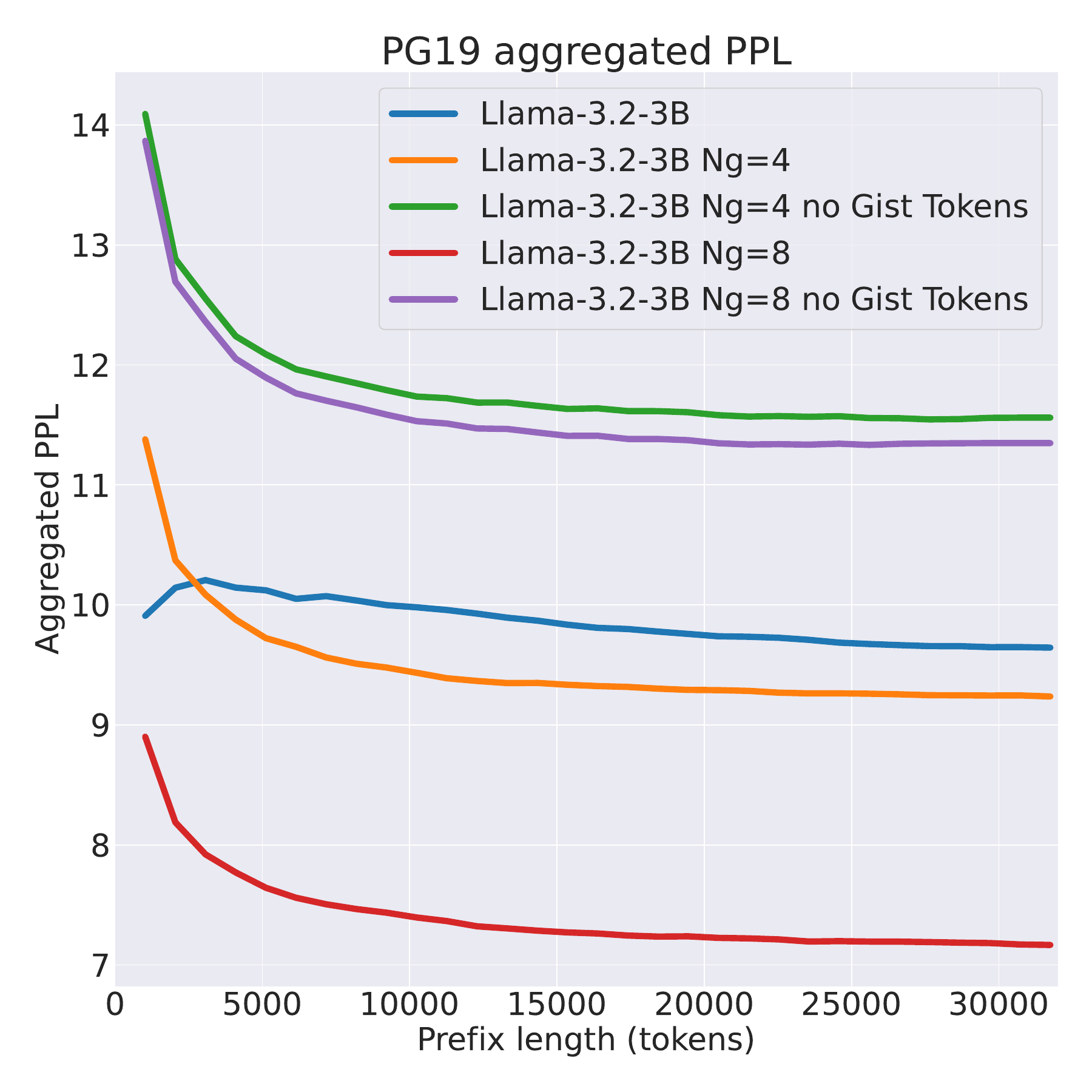}
    \caption{PG19 perplexity for Sentence Llama3.2-3B ($N_g \in \{4,8\}$) compared to the base model across different prefix lengths. The ``no Gist Tokens'' curves represent perplexity calculated while excluding all but the final gist token in each segment.}
    \label{fig:pg_19_log_perplexity}
\end{figure}

\section{Benchmark Compression rates}\label{sec:app_bench_compression_rates}

Table~\ref{tab:short-benchmark-compression-rates} and table~\ref{tab:long-benchmark-compression-rates} shows compression rates on short benchmarks and HELMET (tiny) benchmarks correspondingly. The highest compression rate showed WinoGrande while the lowest compression rates is for synthetic ICL benchmark. For $N_g=8$ model on ICL bench sequence length even gets longer than t he original one. For this table comression rate ($R_c$) was computed by this formula:

\begin{equation}
R_c =  n_{regular} / n_{gist},
\end{equation}

\noindent where $n_{regular}$ is a number of regular tokens, $n_{gist}$ is a number of all gist tokens in processed sequence.

\begin{table}[h!]
\centering
\begin{tabular}{lrrrl}
\hline
 $N_g$ &     ARC &   HS &   MMLU & WG   \\
\hline
 1  & 19.86 &       16.55 &        21.31 &        82.96 \\
 2  &  9.93 &        8.28 &        10.66 &        41.48 \\
 4  &  4.96 &        4.14 &         5.33 &        20.74 \\
 8  &  2.48 &        2.07 &         2.66 &        10.37 \\ \hline
\end{tabular}
\caption{Short benchmarks compression rates.}
\label{tab:short-benchmark-compression-rates}
\end{table}

\begin{table}[h!]
\centering
\begin{tabular}{lrrrrrr}
\hline
 $N_g$  &    cite &     icl &   longqa &   recall &   rerank  \\
\hline
 1  &  32.77 &  7.01 &    31.31 &    17.75 &    25.15  \\
 2  &  16.39 &  3.51 &    15.66 &     8.88 &    12.57  \\
 4  &   8.19 &  1.75 &     7.83 &     4.44 &     6.29  \\
 8  &   4.10 &  0.88 &     3.91 &     2.22 &     3.14  \\
\hline
\end{tabular}
\caption{Long benchmarks compression rates.}
\label{tab:long-benchmark-compression-rates}
\end{table}

\end{document}